\documentclass{article}

\usepackage{arxiv}

\usepackage[utf8]{inputenc} 
\usepackage[T1]{fontenc}    
\usepackage{hyperref}       
\usepackage{url}            
\usepackage[dvipsnames]{xcolor}

\usepackage{booktabs}       
\usepackage{amsfonts}       
\usepackage{nicefrac}       
\usepackage{microtype}      
\usepackage{lipsum}
\usepackage{graphicx}
\usepackage{subfig}
\usepackage[bottom]{footmisc}
\usepackage{hanging}
\graphicspath{ {./images/} }

\title{Classification of Pathological and Normal Gait: A Survey}

\author{
  Ryan C. Saxe\\
  RiskEcon\textsuperscript{®} Lab for Decision Metrics\\
  Courant Institute of Mathematical Sciences\\
  New York, NY 10012 \\
  \texttt{rcs411@nyu.edu} \\
   \And
Samantha Kappagoda\\
  RiskEcon\textsuperscript{®} Lab for Decision Metrics\\
  Courant Institute of Mathematical Sciences\\
  New York, NY 10012 \\
  \texttt{kappagoda@cims.nyu.edu} \\
   \And
  David K.A. Mordecai\\
  RiskEcon\textsuperscript{®} Lab for Decision Metrics\\
  Courant Institute of Mathematical Sciences\\
  New York, NY 10012 \\
  \texttt{mordecai@cims.nyu.edu} \\
}

\begin{document}

\maketitle

For inquiry regarding this paper, please email Ryan Saxe, the corresponding author, at rcs411@nyu.edu and copy the co-authors at riskeconlab@cims.nyu.edu.

\begin{abstract}
Gait recognition is a term commonly referred to as an identification problem within the Computer Science field. There are a variety of methods and models capable of identifying an individual based on their pattern of ambulatory locomotion. By surveying the current literature on gait recognition, this paper seeks to identify appropriate metrics, devices, and algorithms for collecting and analyzing data regarding patterns and modes of ambulatory movement across individuals. Furthermore, this survey seeks to motivate interest in a broader scope of longitudinal analysis regarding the perturbations in gait across states (i.e. physiological, emotive, and/or cognitive states). More broadly, inferences to normal versus pathological gait patterns can be attributed, based on both longitudinal and non-longitudinal forms of classification. This may indicate promising research directions and experimental designs, such as creating algorithmic metrics for the quantification of fatigue, or models for forecasting episodic disorders. Furthermore, in conjunction with other measurements of physiological and environmental conditions, pathological gait classification might be applicable to inference for syndromic surveillance of infectious disease states or cognitive impairment. 
\end{abstract}


\section{Introduction}
\par
Gait recognition is a term commonly referred to as an identification problem within the Computer Science field. Empirical methods (e.g. data, algorithms, models) are employed to identify an individual based on their pattern of ambulatory locomotion\footnote{This paper specifically focuses on gait in regards to human bipedal movement. Gait is still a relevant metric for both animal locomotion and human quadrupedal locomotion (e.g. crawling), but that is outside of the scope of this paper.}. Humans are complex, and the amalgamation of variables that yield their corresponding, quasi-unique, gait can be analyzed and learned. Research applying gait recognition methods to relevant data may have potential to yield empirical indicators with explanatory power for a variety of physiological, emotive and/or cognitive conditions. With regard to such empirical indicators as syndromic diagnostic signals, gait perturbation may be a proxy for attributing such perturbations to underlying pathological state transitions. The focus of this paper is on the application of signal processing to measurable state transitions in the context of pathological gait diagnostics. 

This paper reviews how such empirical indicators can be modeled and learned in order to classify subjects with regards to either normal\footnote{This paper defines “normal gait”, according to Whittle (2002), as a standardized model of gait in order to assess the gait of patients. This means “normal” is defined differently depending on the expected patient per study (e.g. A study on elderly women will have a different definition of normal versus a study on young men).} or pathological\footnote{This paper defines “pathological gait”, according to Whittle (2002), as gait affected by some known physical or neurological condition in a manner such that it deters from “normal gait”. 
} gait with either longitudinal or non-longitudinal classification methods. “Longitudinal classification” is employed to assign individual gait to a class or category, based on a specified cross-sectional variable observed over time, whereas “non-longitudinal” classification uses cross-sectional methods to assign individuals into categories based upon a non-temporal target variable\footnote{A target variable is a feature within the data that determines the classes/categories for classification.}. The more commonly studied gait recognition problem of identifying an individual based on their signature gait is an example of a normal gait, non-longitudinal classification problem. \par
\textbf{Normal Gait}: There are a diversity of features that may be utilized to classify individuals without pathological processes related to gait including stride length, frequency of steps per distance (velocity), pattern of regularity of steps. These features can be used to classify individuals related to several conditions or states of adaptive functionality such as fatigue, performance readiness, fitness, motor coordination, etc. Once such accurate and reliable classification is achieved non-longitudinally, it can be further explored to identify individuals related to a meaningful target variable in the future as long as the data is longitudinal. This type of classification can have value to specific populations such as athletes by forecasting performance or injury.
\par
\textbf{Pathological Gait}: The diversity of features described to classify individuals with normal gait may also be used to detect neuropathological processes such as in patients with neurologic (e,g. Parkinson’s disease) or psychiatric (e.g. mania, depression) disorders. Symptoms of psychomotor retardation or agitation are considered to be well documented indicators of a variety of neurologic or psychiatric disorders (Bennabi, Vandel, Papaxanthis, Pozzo e Haffen, 2013). This may yield cross-sectional empirical results for classifying or diagnosing neurological or psychiatric disorders, and may also facilitate longitudinal classification using data sets that exhibit explanatory or predictive power regarding target variables at some future time, relative to the time when the gait features were measured. Such longitudinal classification might improve analyses related to a patient’s prognosis, response to treatment, or leading indicators relevant to health- and safety- related outcomes such as work absences, psychiatric hospitalizations, behavioral incidents associated with affective and cognitive impairment (e.g. dementia, delirium), and emergency room incidents precipitated by fatigue conditions associated with inflammatory response to infectious agents.
\par
However, prior to pursuing these potential research applications, it is important to obtain a reasonably thorough understanding of the underlying methods and tools required for gait recognition. The methods of gait recognition can be divided into multiple steps. First and foremost, there is data collection, which is created either from video surveillance or through biometric devices such as accelerometers and force-sensors. Although a variety of publicly available datasets containing such data have been identified (as surveyed in Section 2), the sparsity of these datasets is noteworthy. Subsequent to data collection, two stages of data transformation are typically applied. Section 3 describes the initial pre-processing techniques. This section describes the differences between pre-processing video-data and biometric data as well as two different approaches to feature extraction in regards to gait data: model-based and model-free. Model-based pre-processing employs physiological, anatomical, and structural fundamentals regarding the dynamics of the body in order to explicitly specify a model. The model-free approach, on the other hand, employs no such prior knowledge (Shirke, Pawa, Shah, 2014). The last step of pre-processing data is dimensionality reduction, which is a form of feature extraction intended to reduce the dimensionality of the data in order to remove redundancy and noise. The popular reduction techniques of Principal Component Analysis (PCA) and Linear Discriminant Analysis (LDA) are surveyed at the end of Section 3. 
\par
Once the data is processed, it can be used for classification. The commonly used classification techniques in the field of gait recognition are K-Nearest Neighbors (KNN), Support Vector Machines (SVM), Hidden Markov Models (HMM), and Neural Networks (NN). Dynamic Time Warping (DTW) is a commonly applied metric to quantify distance between time series, and can be used for both classification and preprocessing. Section 4 of this paper will discuss these classification methods as applied to all permutations of normal vs. pathological gait and longitudinal vs. non-longitudinal classification for the problem of gait recognition. This is then followed in Section 5 by potential future projects and applications in the field, and concluded in Section 6.

\section{Data Survey}
\label{sec:data}
There are a variety of data sets currently available to conduct research within gait recognition applied to human bipedal locomotion. Several data sets that represent examples of available data are reviewed in this section according to their potential application for research on normal or pathological gait implementing either longitudinal or non-longitudinal classification. Table one summarizes these data sets according to these categories.

\subsection{Normal Gait and Non-Longitudinal Data}

\subsubsection{CMU MoBo}

This dataset was generated by the Robotics Institute at Carnegie Mellon University. It consists of 25 individuals walking on a treadmill in four different ways (slow, fast, with incline, while holding a ball), captured by cameras from six different angles, as well as providing silhouette extracted images for each permutation. For specifics on requesting this database, and particular information regarding the data such as sequence length and camera specifics, refer to (Gross, Shi, 2001).

\subsubsection{USF Gait Baseline}

This is a large dataset, over a terabyte of video data, generated by the University of Southern Florida. It consists of 1870 sequences from 122 participants who walked on a lawn under different circumstances. Nearly each participant walked with all the permutations represented in Figure 1.

\begin{figure*}[b]
  \centering
  \includegraphics[scale=0.5]{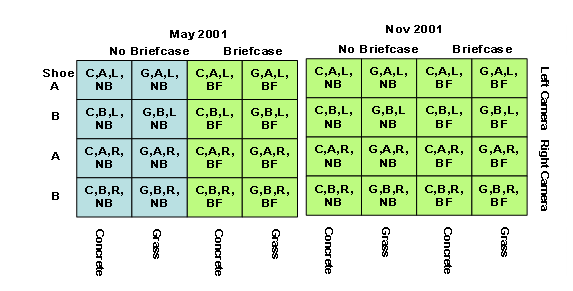}
  \caption{(S. Sarkar et al, 2005): All permutations of the USF Gait Baseline data.}
  \label{fig:fig1}
\end{figure*}

\subsubsection{CASIA}

This database is provided by The Institute of Automation, Chinese Academy of Sciences (CASIA). It consist of three\footnote{There is a fourth dataset according to the CASIA website, however it does not appear to be readily available.} datasets (A,B,C), each of which consists of video and silhouette footage of participants walking at a variety of angles, speeds, and wearing items such as backpacks and coats. Dataset A is the smallest, with 20 participants. Each participant has twelve sequences, four for each direction (0, 45, and 90 degrees) with respect to the camera. Dataset B is substantially more meticulous, as it consists of 124 subjects, each of which has sequences corresponding to eleven different angles with respect to the camera. Furthermore, subjects are monitored normally, as well as while wearing a coat or carrying a bag. Dataset C consists of 154 participants under four different walking conditions (slow, fast, normal, and with a bag). However the novelty of this dataset is that the video-footage was taken at night with an infrared camera. 
\par
It should be noted that the silhouette extracted image sequences are publicly available for download, however the full 10GB database must be requested.

\subsubsection{Physiobank: Long Term Dynamics}

\textit{PhysioBank}\footnote{The list of available PhysioBank databases related to gait can be found \href{https://physionet.org/physiobank/database/\#gait}{here}.}
 is a collection of physiological databases, and happens to contain a variety of databases and studies in regard to gait applied to biometric time series data. All of the data is publicly available, but some of the files are compressed by PhysioBank and can only be read with their specified, publicly available, software. This applies to both this dataset and all following PhysioBank datasets.
\par
This dataset consists of force-sensor data of ten young men (average age 21.7) generated by placing a force-sensor in the shoe of the participant. Each participant has six distinct time-series data files, each one hour long. The difference between each file is the pace (slow, normal, or fast) and whether or not their stride was coordinated via metronome. 

\subsection{Pathological Gait and Non-Longitudinal Data}

\subsubsection{PhysioBank: Neurodegenerative Disease}

This appears to be the most extensive databases available on PhysioBank. There are 64 participants that are grouped in terms of multiple neurodegenerative diseases, as well as a control group. The goal is to focus on stride and variation. Figure 2 demonstrates stride interval over time in subjects without neurologic disease and in patients with three categories of neurologic diseases (Parkinson's disease, Huntington's disease, Amyotrophic Lateral Sclerosis). A visual inspection of the patterns illustrated in this figure indicates that there may be a stride interval ‘signature’ for a neurological disease, which is distinct from normal gait. The data consists of time series data (displayed below in Figure 3) as well as signal data on both the left and right foot taken from a sensor.

\begin{figure*}[!b]
  \centering
  \begin{minipage}[b]{0.48\textwidth}
    \includegraphics[width=\textwidth]{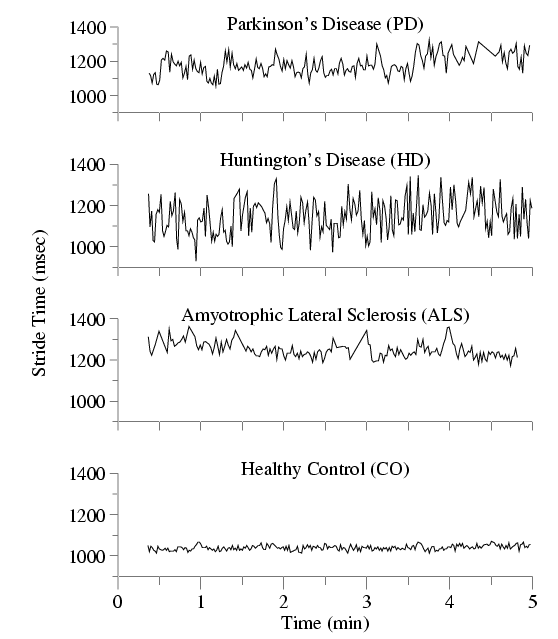}
    \caption{(Goldberger et al, 2000): Neurodegenerative disease timeseries data.}
  \end{minipage}
  \hfill
  \begin{minipage}[b]{0.48\textwidth}
    \includegraphics[width=\textwidth]{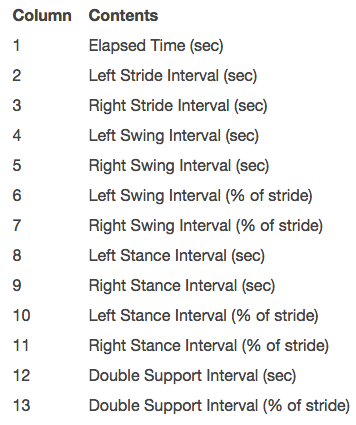}
    \caption{(Goldberger et al, 2000): Columns corresponding to  neurodegenerative disease data.}
  \end{minipage}
\end{figure*}

\subsubsection{Physiobank: Parkinson's Disease}

This specific disease is often studied in tandem with gait because of how much gait is affected by Parkinson’s Disease. This dataset consists of patients with (93) and without (73) the disorder, and their gaits during a two-minute walk. Demographic data is also available for each patient. Furthermore, the data is generated from 8 sensors under each foot of each patient! So this specific biometric is similar to that of a force plate.

\subsection{Normal Gait and Longitudinal Data}

\subsubsection{Physiobank: Long Term Movement}

This database contains two sets of three-dimensional accelerometry data from 71 elderly persons. One of the data sets is recorded over three days while at home, while the other is obtained from a conducted in-lab minute-long walk. Finally, each participant has metadata such as age, gender, number of falls in the past year, clinical fall-risk assessment scores.

\subsubsection{Physiobank: Tai Chi}

This study collected data on 87 adults that had no prior Tai Chi experience versus those with at least five years (referred to as experts). Foot switches were used as well as an acquisition monitor along the waist in order to collect the data over ten minutes of walking. Tai Chi experts were only measured once, while those inexperienced — either the control group or the group learning Tai Chi — were monitored several times during the period of the study.

\subsubsection{Kaggle Competition for Gait Recognition}

Kaggle.com, a website dedicated to data science challenges and competitions, had a competition for gait recognition based on phone data: accelerometry and time for 387 participants, combining to a total of 30-million samples. The time is represented as the number of milliseconds since January 1st, 1970 at midnight UTC\footnote{It should be noted that to avoid leakage in this competition, some of the variables, including the time variable, were altered and hence is not consistent across subjects. The discussion on this can be found \href{https://www.kaggle.com/c/accelerometer-biometric-competition/discussion/5721}{here}}. The accelerometry data is represented by each axis (x,y,z) in terms of gravitational force. For example, a phone set on the table facing upwards would be represented by the point (0,0,-9.8). 

\subsection{Pathological Gait and Longitudinal Data}

\subsubsection{Kaggle Competition for Parkinson’s Disease}

Similar to the Kaggle competition referred to in Section 2.3.3, this competition was based on phone data. Additionally, this dataset has a control group and a group of patients with Parkinson's disease, each consisting of eight participants. Other additional information, such as GPS data and metadata was available for each participant.

\begin{table*}[b]
  \centering
  \includegraphics[scale=0.5]{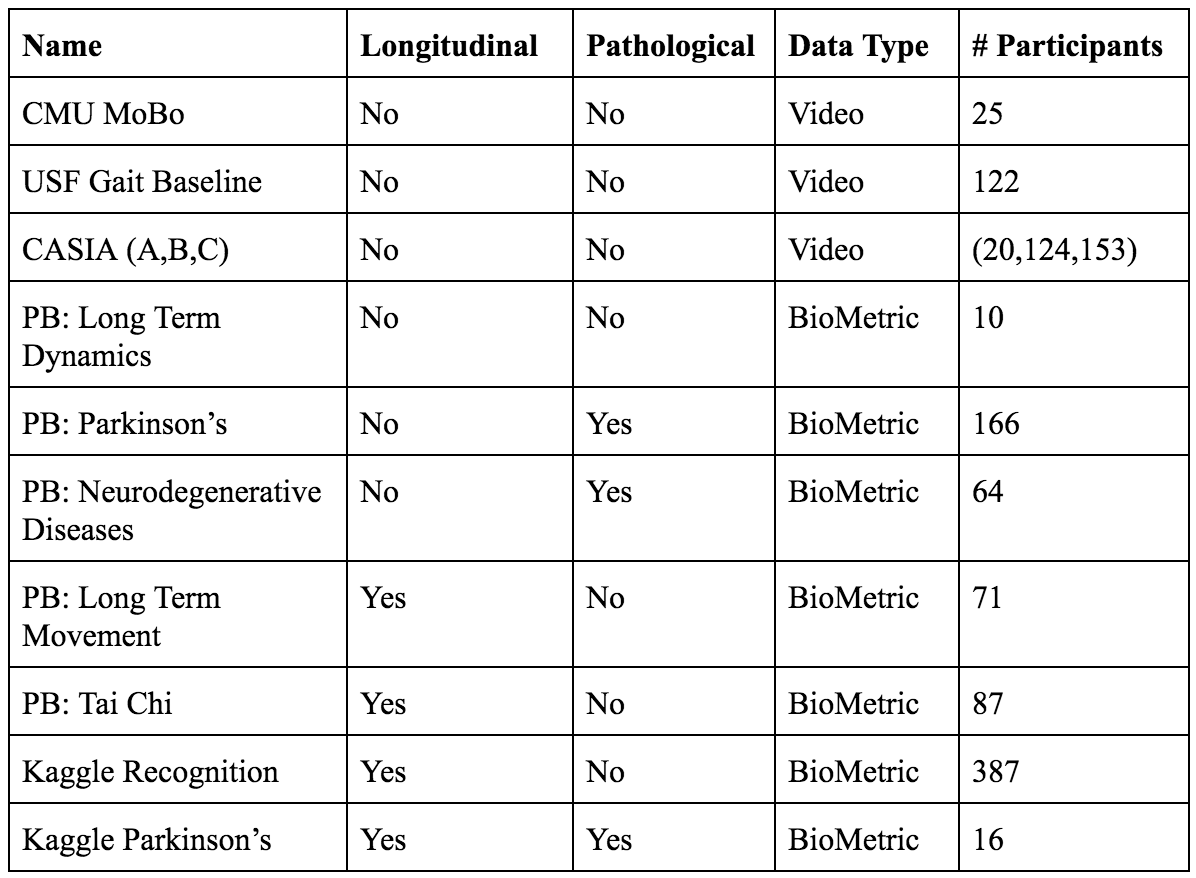}
  \caption{Descriptions of reviewed datasets for research on gait recognition.}
  \label{fig:table1}
\end{table*}

\section{Data Preprocessing}

For the problem of gait recognition, the data must be cleaned first to remove extraneous information and noise. Then, once the data has been properly cleaned, it should go through two forms of feature extraction. First, additional relevant information and biometrics such as cadence — number of steps per minute — are extracted. Once the additional features are added to the data, dimensionality reduction based feature extraction techniques should be employed in order to prepare the data for classification. 

\subsection{Noise Removal}

Every dataset has noise, which will impair the performance of any classification method under most circumstances.The following sections review noise removal techniques for both biometric and video data.

\subsubsection{Biometric Data}

Biometric data often consists of both accelerometry/force-sensor data and patient metadata. Metadata can undergo feature selection, whereby a subset of data columns are chosen that best represent the study in question. This is commonly implemented in four ways. First, features with a density of null values are either removed or filled in by some convention (i.e. average value for non-null values of that feature). Second, features with too high or too low variance are unlikely to influence the classification results, and they can be removed\footnote{Entropy and Information Gain are common tools for evaluated this variance both for removing insignificant features as well as prioritizing features with potentially substantial predictive power.}. Third, a domain expert for the study can select the relevance of a feature, or lack thereof, to the study. Lastly, the relevance of features can be computed and analyzed via feature importance algorithms \footnote{The specificity of these algorithms are outside of the scope of this paper. If there is further interest in feature importance methodology, Tang, Alelyani, and Liu (2014) provide a sufficient survey.}.
\par
Accelerometer and force-sensor data, however, require more nuanced techniques. This is because the noise within this data is often a specific sequencing that is irrelevant to gait, or a malfunction in the data-collection device. Consider accelerometry data generated by a cellular device. Suppose that a participant dropped the device. This can be detected by a jolt in velocity in the downward direction followed by a sudden stop and vibration upon collision with the floor. This type of biometric data can be cleaned by algorithms\footnote{Many of these algorithms fall into the category of anomaly detection, which is a topic outside of the scope of this paper.Further information on these techniques are surveyed by Chandola, Banjeree, and Kumar (2009)} that detect movement unrelated to gait such as extreme movement or no movement at all. 

\subsubsection{Video Data}

Silhouette extraction is the initial step to any processing of video data for gait recognition. An image or video has a significant amount of extraneous information, and hence it is crucial to extract the focal point of the data (i.e. the individual moving). This is an important application of dimensionality reduction as it effectively removes any and all features present in the data that are not directly related to the movement of the individual. \par
Removal of the background\footnote{There are a variety of different algorithms used for background subtraction that will not be surveyed in this paper. A subset can be found in (Benezeth, Jodoin, Emile, Laurent, Rosenberg, 2010).} of a video-feed works under the assumption that there is little-to-no motion in the video outside of the individual whose gait is in question. Since a video is a sequence of images with the same dimensions, the pixels relevant to the background only change drastically when the individual moves within that localized area. (Bouchrika, Nixon, 2007)
\par
Once the background is removed, the image should be blank everywhere other than the desired silhouette. Sometimes this is not the case, and additional noise persists. Morphological processing is a technique for processing geometric structures that is used to isolate that larger anatomical structure: the goal for extraction (Shaikh, Saeed, Chaki, 2014). This is computed by examining the full image with a structure, often a small square. This structure is placed in every possible position of the image, and records the density of blank pixels in each area. This analysis can then remove noise isolated from the target silhouette. 

\begin{figure*}[!b]
  \centering
  \begin{minipage}[b]{0.7\textwidth}
    \includegraphics[width=\textwidth]{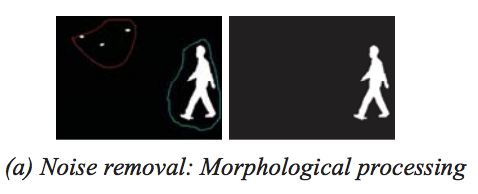}
    \caption{(Shaikh, Saeed, Chaki, 2014): Image representing a silhouette pre- and post- morphological processing.}
  \end{minipage}
  \hfill
  \begin{minipage}[b]{0.7\textwidth}
    \includegraphics[width=\textwidth]{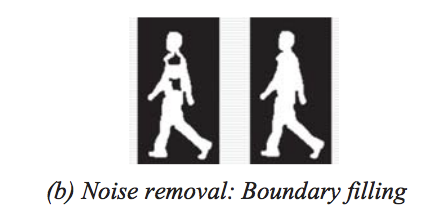}
    \caption{(Shaikh, Saeed, Chaki, 2014): Image representing a silhouette pre- and post- boundary filling.}
  \end{minipage}
\end{figure*}

\par
Finally, the silhouette extracted from the image can have gaps as seen in Figure 6. These must be filled in order to properly represent the individual in the data. It is important not to fill in the space between the arm and the body, yet it is important to fill in a hole that doesn’t properly represent anatomy. A heuristic solution to this problem is to search for holes completely encompassed by the silhouette, which turns out to be a sufficient solution (Shaikh, Saeed, Chaki, 2014).
\par
Once there is a proper represented silhouette for the image in question, model-free or model-based approaches can be applied to extract features. For example, one can calculate stride or gait-cycles\footnote{This paper defines “gait cycle”, according to Whittle (2002), as the movement and time beginning from initial contact until the state of initial contact is repeated. Whittle gives the example of initial contact with the right foot. This is the beginning of the cycle. The subject then steps with their left foot, and then their right foot again. When the right foot contacts the floor for the second time, the initial contact has repeated and this is hence the end of the cycle.
} based on the contour-bounding-box around the silhouette (Shaikh, Saeed, Chaki, 2014). 

\subsection{Feature Extraction}

Whether the data is biometric or video, it is still describing the same concept: an individual’s gait. And there are plenty of potential features relevant to gait, such as stride, that can be extracted from silhouettes, accelerometry data, force-sensor data, and other structural and functional data (e.g. skeletal extraction)\footnote{By way of illustration and for further background on feature engineering implementing skeletal extraction, see (Saxe, 2018).}. The different kinds of feature extraction methods for gait fall into two categories: model-based and model-free.

\subsubsection{Model Based}

The model-based approach is centered around understanding anatomical dynamics. Concepts and biometrics based upon this understanding can be extracted from data in order to identify and quantify gait. There are a variety of different metrics, all of which require knowledge of dynamics and anatomy, that can be applied here. Some of the more common model-based metrics are joint angles and anatomical segmentation.
\par
While model-based approaches can achieve high performance due to these complex metrics, implementation can be impractical. In order to extract such clear information on the dynamics of the body in a video, the quality of video-data must be high and this is often very difficult to obtain in real life applications (Shirk, Pawa, Shah, 2014) (Bouchrika, Nixon, 2007). Consider the application of gait recognition through security footage. This type of footage is not typically of high quality, and hence a model-based approach may not be able to identify a perpetrator accurately. Additionally, given the considerable requirement of quality for initial information as applied to the processing and algorithms, the computational cost is also high (Shirk, Pawa, Shah, 2014) (Bouchrika, Nixon, 2007). 
\par
Overall, this approach appears to have a higher bar for both use and implementation. And hence an application that is intended to be widely applicable may need to use a model-free approach, as it bears less of a computational cost (Shirk, Pawa, Shah, 2014). 

\subsubsection{Model Free}

Unlike the model-based approach, the model-free approach extracts a variety of gait-relevant features from the data without any prerequisite anatomical knowledge. While there are a plethora of features, such as joint angle, that cannot be extracted in this manner, there is a substantial amount that can be extracted. A common model-free approach is to compute the Gait Energy Image (GEI), which creates a two-dimensional representation of an individual’s gait from a sequence of silhouette images (Man, Bhanu, 2006). A simple example applied to accelerometry data is conceptualizing a potential score based on overall movement discerned from the data. An example of such score is the Activity Index (Bai et al, 2016), which is a quantification of the activity level of the individual. 
\par
The weakness of the model-free approach is also the strength of the model-based approach. With this lack-of-knowledge and potential variation within data, the model-free approach is less robust, and hence requires tailoring to specific data (Shirk, Pawa, Shah, 2014). This creates an issue in scalability for larger applications when it comes to image data, as the representation from cameras, lighting, etc. could yield different results with the same model-free analysis.

\subsection{Dimensionality Reduction}

Prior to classification it is important to reduce the dimensionality of the dataset in order to both maximize performance and reduce overfitting\footnote{Although this importance varies depending on the robustness of the classifier and data structure.}. Dimensionality reduction is a form of feature extraction, but instead of generating additional features from the given data, an entirely new feature space is created. Given that individual-level gait data has such a large range of variables to look at — especially when considering video-data — it is necessary to discern a set of features that best describes the data. While there are a variety of sufficient methods capable of reducing the dimensionality of the data, two commonly applied algorithms to gait recognition are Principal Component Analysis and Linear Discriminant Analysis.

\subsubsection{Principal Component Analysis (PCA)}

PCA is an algorithm used for dimensionality reduction and feature extraction. PCA achieves this by finding k principal components of the data, reducing the n-dimensional data matrix to k-dimensional\footnote{Where k is substantially less than n, often noted mathematically as k << n.}. Principal components are linearly independent vectors that form an orthonormal basis for the new k-dimensional space. Conceptually speaking, this new k-dimensional matrix is a transformation into a new topological space that maximizes the variance between features in this space due to the nature of these components.

Let $M$ be an n-dimensional data matrix. First, PCA requires discovering the k orthonormal axes, represented by $w_i = \forall i \in [1,k]$ that correspond to the eigenvectors of the k-largest eigenvalues of the covariance matrix.

$$\sum_{x \in M} \frac{(x - \widetilde{x})(x - \widetilde{x})^T}{n}$$

Hence any row of n-dimensional data can now be transformed into the concept space by finding its components by computing $W^T (x - \widetilde{x})$. This will maximize the variance within the features of this space because each feature is selected based on this new orthonormal basis. Then, the entire matrix  can be mapped into this new space, and handle our reduced data for the problem at hand: gait recognition. Of note, there are variants of PCA (e.g. Kernel Principal Component Analysis (KPCA), Sparse Principal Component Analysis (SPCA)) that can be better suited to solve gait recognition problems (Ekinci, Aykut, 2007).

\subsubsection{Linear Discriminant Analysis (LDA)}

LDA, like PCA, is a linear transformation technique for taking the n-dimensional feature space and reducing its dimensionality. However, rather than maximizing the variance between the features, LDA attempts to maximize the variance between the classes. Under the assumption of a gaussian distribution and similar variance across variables, LDA can analyze the variance across the data matrix $M$ by looking at the covariance matrix with a representative vector for each class. Let $\mu_i$ be the mean corresponding to the $i^{th}$ class.

$$\sum_{i = 1}^k \frac{(\mu_i - \widetilde{\mu_i})(\mu_i - \widetilde{\mu_i})^T}{k}$$

And just like PCA, the eigenvectors corresponding to the above matrix\footnote{There cannot be more than k eigenvectors, as the matrix has at most rank k.} forms a linear transformation of any data row to a new k-dimensional concept space designed to separate the k-classes.
\par
This optimization towards separation of classes via the maximization of variance within the classes is the key towards the success of LDA in the current literature of gait recognition; if the goal is to identify an individual, the desired result is a separation as described. Furthermore, LDA can be employed as a classification technique using Bayesian probability along with the distribution of classes in the labeled data, which has been applied to gait recognition (Su, Liao, Chen, 2009).

\section{Classification}

As previously stated, the classification problems for gait recognition can be grouped by whether they are longitudinal or not, and such grouping is helpful for organizing practical research questions related to the application of gait recognition methods. However, both groups are mathematically similar as the main distinction between groups is the time period during which the target feature (the class) was measured. Hence, regardless of which permutation of normal vs pathological gait and longitudinal vs non-longitudinal classification, any of the algorithms reviewed in this section can be applied to solve the problem, as long as the data and problem in question fit the algorithmic requirements. 

\subsection{Algorithms}

\subsubsection{K-Nearest Neighbors (KNN)}

\begin{figure*}[!b]
  \centering
  \includegraphics[scale=0.3]{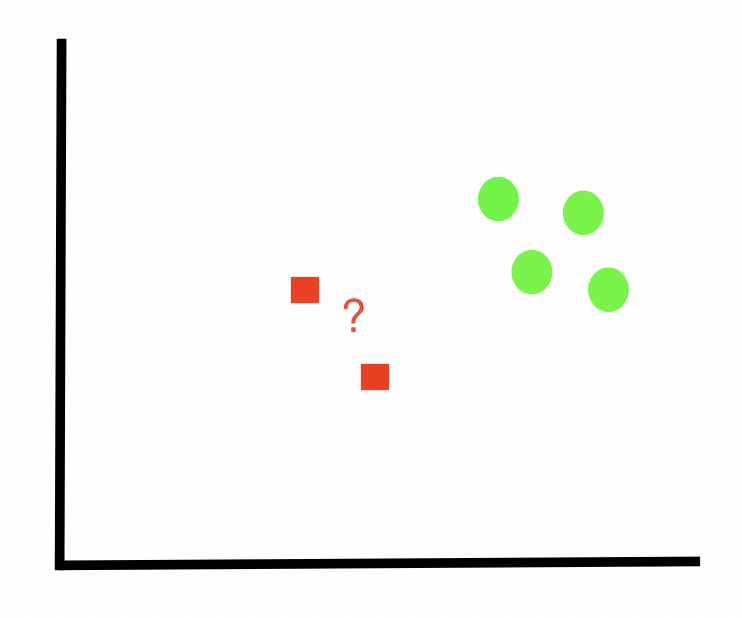}
  \caption{Goal: classify the question mark. Is it more likely to be a red square or a green circle?}
  \label{fig:knn}
\end{figure*}

K-Nearest Neighbors is a classification algorithm that given any data point x — for gait recognition this is the representation of an individual’s gait — and some number k, x is classified by a majority rule based on the known classification of the k-nearest data points in the training set. In order to find the k-nearest data points, there needs to be an understanding of distance within the feature space, which will be covered in section 4.2. 
\par
K-Nearest Neighbors can be improved upon by weighting neighbors based on this distance. Consider Figure 7 as a representation of the data within the feature space.

KNN with k = 5 would classify this mark as a green circle, even though it is substantially more similar to a red square. However, if each data point is weighted with a value based on the distance to the question mark rather than an equal weight of 1/k, then the resulting classification would be a red square.  

While KNN is not the most robust algorithm, it is simple to implement, and useful. Its most common application in the gait recognition literature is as a benchmark for performance study — In testing an algorithm it is quite common to compare performance to that of a basic KNN model given how easy KNN is to implement.

\subsubsection{Support Vector Machines (SVM)}

Support Vector Machines\footnote{For a deep overview of how SVMs work and the theory behind them, please refer to Burges’ tutorial (1998).}(Boser et al, 1992) (Cortes, Vapnik, 1995) are supervised learning algorithms successfully applied in many fields, including gait recognition. The objective of this algorithm is to discover a hyperplane of lower dimensionality that can separate data-points by the largest margin in order to classify them. 

Let our data points be in the form $(x_i,y_i)$ where $x_i$ is the feature vector, and $y_i$ is the binary classification associated with $x_i$. Assuming that the data in question is linearly separable\footnote{Data is considered linearly separable if there exists some hyperplane that divides the data such that the binary classifier is capable of perfect classification.}, the following must be true for the calculated support vector $w$ where $b$ is a bias involved for representation of the hyperplane.

$$(w \cdot x_i + b)y_i \geq 1 \forall i$$

However, the assumption of linear separability is too strong. To solve this problem the margins found can be softened\footnote{Softened here is defined as allowing for misclassification.} by introducing a hinge-loss function. This creates the following objective to the optimization problem of finding the $w$ and $b$ that generate a hyperplane that minimizes misclassification in the training data.

$$min_{w,b}\sum_j max\{0,1-(w \cdot x_j + b)y_j\}$$

More importantly, this binary classification problem can be extended to a multi-classification problem either using one-vs-rest or one-vs-one methods, but in the literature of gait, one-vs-rest is the most common implementation due to computational efficiency. One-vs-rest functions on the axis that the support vector\footnote{A support vector here is a binary classifier described by the math above. } that yields the largest output is the proper classification.

The last aspect of SVMs is called a kernel, and is a large contributor to why SVMs are so robust. A kernel k is a similarity metric with the following restriction: Let $k: A \times A \rightarrow {\rm I\!R}$, then $\exists f: A \rightarrow B$ such that $\forall x,y \in A \quad k(x,y) = <f(x),f(y)>$

Let’s assume that there are $c$ unique classes and $n$ training samples where the i\textsuperscript{th} training sample is written $(x_i,y_i)$ and the j\textsuperscript{th} class is written $C_j$. Let $z_{ij}$ be the calculated weight of the i\textsuperscript{th} training sample for $C_j$. And let $y_{ij}$ be 1 if $y_i = j$ and 0 otherwise. Then any unlabeled data-point $x$ can be classified as the corresponding class $C_j$ to the maximal sum in the following equation:

$$max_{j=1}^c \sum_{i=1}^n y_{ij} * z_{ij} * k(x_i,x^{\prime})$$

In general, given how robust SVMs are, with the proper data processing and kernel selection, they will be a reasonable approach to almost any gait recognition problem, and hence this classification algorithm is quite popular in the literature.

\subsubsection{Hidden Markov Models (HMM)}

A Hidden Markov Model, theorized by Leonard Baum and his colleagues (Baum et al, 1970), is a statistical model that can be used for learning. This model is built upon the assumption that the system it is modeling is a Markov process, but with hidden states where each state is only dependent on the previous state. The goal of this model is to take an input observation sequence, and discern which sequence of states is most likely to have produced this observation. We can more formally define these concepts as follows:

Let $S = (s_1,s_2,\ldots,s_n)$ be the set of possible states and $O = (o_1,o_2,\ldots,o_m)$ be the set of possible observations.

Then we can model an HMM, called $M$, where $M = (A,B,x)$. $A$ is the transition matrix defined such that $A[i,j]$ is the probability that, in a sequence of states, $s_j$ follows $s_i$. $B$ is a matrix defined such that $B[i,j$ is the probability that an observation $o_i$ is produced from the state $s_j$. And $x$ is an array defining the probability of initial states such that $x_i$ is the probability that the first hidden state is $s_i$.

Given the problem of gait recognition, we are trying to model the gait of each specific individual. Let $M_k$ be the model that best represents the k\textsuperscript{th} subject for classification. Then the following function can be used to classify an observed sequence $O$ of an unknown subject:

$$max_{\forall k}P(O|M_k)$$

In order to apply this to gait recognition, we must have a method for creating all models $M_k$, as well as an algorithm for evaluating the probability that an observed gait belongs to any such models. Evaluation of this probability for HMMs is usually computed through the Viterbi algorithm, and the matrices A, B, and x can be updated via the Baum-Welch algorithm.

As far as creating the proper HMM for any given individual, the literature suggests the best approach is by modeling one gait cycle. When walking, a person takes a step with one foot and then the other. This can then be represented in terms of N states, where the initial state would also follow the Nth state, as these states are cyclical (Hai, Thuc, 2015). This is under the assumption that each state can be represented by a part of a temporal gait sequence.

\subsubsection{Neural Networks (NN)}

Neural Networks (NN) are a powerful\footnote{The Universal Approximation Theorem stated by Csáji (2001) illustrates that by a specific structure a NN can approximate any continuous function. Given that the solution to many machine learning problems is function approximation, this is a very important claim.} set of machine learning algorithms. These algorithms follow a specific structure built of layers of neurons, where neurons take some value\footnote{In the literature, this value is called the “activation”. } and each neuron in one layer is connected to all neurons in the succeeding layer. The first layer is the input layer. This layer has a neuron to correspond to each considered feature such that the data-vector can be fed into the network. Following the input layer is a sequence of one or more layers, called hidden layers. 

Let $L_i$ be the i\textsuperscript{th} hidden layer and $L_{i,j}$ be the value of the j\textsuperscript{th} neuron in that layer. During the training portion of the network\footnote{The most common way to train a NN is using backpropagation optimized with an algorithm like Stochastic Gradient Descent or Adam. Backpropogation was developed in the 1970s, and Rumelhart, Hinton, and Williams (1986) were the first to display its potential for NNs. The details of the algorithm, however, are out of the scope of this paper in terms of necessity of understanding for implementation of Gait Recognition.}, each neuron in the next layer will learn a weight corresponding to that specific neuron. Let $w_{k,j}$ be the weight for the k\textsuperscript{th} neuron in $L_{i+1}$ corresponding to $L_{i,j}$. Then the activation of this neuron can be described as follows:

$$L_{i+1,k} = F(\sum_{j=1}^{|L_i|}L_{i,j} * w_{k,j})$$

Where F is some chosen activation function\footnote{It is often best for the values of neurons to be between one and zero, or one and negative one. An activation function keeps the value of neurons between a desired range in a mathematically sound way in terms of the structure of NNs. Many of these functions were reviewed and analyzed by Karlik and Olgac (2011).}. The data flows through the hidden layers in the manner described until it hits the last layer, the output layer. This layer is what determines classification. Let’s consider the problem of Gait Recognition. If the study has N subjects, then this output layer will consist of N neurons, each corresponding to a specific subject. The softmax activation function is then applied to this output in order to convert it to a probability distribution. And hence The neuron in the output layer with the highest probability would correspond to the proper subject classification of the gait in question.

Furthermore, there are a variety of different types of NNs with additional complexities implemented in their structure. Long-Short Term Memory (LSTM)\footnote{The LSTM architecture is outside of the scope of this paper. For further explanation on its relevance, please refer to the conceptual overview and introduction of the LSTM architecture posted by Chris Olah, \href{https://colah.github.io/posts/2015-08-Understanding-LSTMs/}{"Understanding LSTM Networks"}.} Networks have been proposed for the application of pathological gait classification (e.g. Khokhlova et al. 2019). Another architecture applied to gait recognition applications is the Convolutional Neural Network (CNN)\footnote{A convolutional layer is designed to process the input comparably to how the brain processes visual stimuli by convolving some filter onto the input in order to attain higher specificity within the transformation of the input. While the details of CNNs are out of the scope of this paper, LeCun et al. (1998) is one of the initial applications of learning CNNs (using a novel structure named LeNet-5) for real world problems — document recognition — and can provide insight on the underlying structure.}. As a generally accepted practice, the application of CNNs to image classification is well established (Krizhevsky, Sutskever, Hinton, 2012). The  development by Gadaleta and Rossi (2018) demonstrates the utility of applying CNN in conjunction with SVM to accelerometry-based gait recognition.

\subsection{Dynamic Time Warping (DTW)}

Dynamic Time Warping is a similarity metric employed for comparing temporal sequences\footnote{It should be noted that DTW can also be used as a classification technique as seen in (Switonski, Michalczuk, Josinski, Polanski, Wojciechowski, 2012), however this paper will focus on its application as a similarity metric.}. More specifically, DTW takes two temporal sequences that follow different time frames and align them for comparison. This can be applied in the field of gait recognition because it enables quantifying similarity between gaits — and their cycles — without worrying about precise time measurements and alignment. DTW achieves this by finding a path, called the warping path, that aligns the two temporal sequences and assesses the cost of such a path. 

Assume two sequences $X = (x_1,x_2,\ldots,x_n)$ and $Y = (y_1,y_2,\ldots,y_m)$. Let $D$ be an $nxm$ matrix representing the potential costs of warping in the following manner:

$$D(i,j) = min\{D(i-1,j),D(i,j-1),D(i-1,j-1)\} + cost(x_i,y_i)$$

Where $cost(x_i,y_i)$ is the cost of aligning parts of the respective sequences. Given this matrix, the DTW algorithm finds the cheapest path using dynamic programming by starting at the end and selecting, through a recursive process, the cheapest path that ends at either $D(n-1,m)$,$D(n,m-1)$, or $D(n-1,m-1)$. The cost of this path is hence the distance between the temporal sequences in question.

Each algorithm covered in Section 4.1 requires, in some way or another, the comparison of feature vectors. Given that gait is a biometric that is described over time, DTW is a similarity metric well suited for this type of comparison. Both KNN and SVM implement DTW in a similar fashion. For KNN it is the given distance function, and for SVM it is the kernel method (Gudmundsson, Runarsson, Sigurdsson, 2008). HMMs require a bit more finesse, as the method used to find the best representative class is the Viterbi algorithm. Fortunately, since HMM is also a path-based algorithm, DTW can be enhanced (EDTW) via HMM by extending the Viterbi algorithm. This is then used as a classifier, and has been shown to outperform DTW when implemented as a classification technique (Kumar, Kant, Shachi, 2013). Lastly, NNs can be implemented such that DTW is employed as a kernel-like function for learning (Iwana, Frinken, Uchida, 2016)

\section{Future Research Directions}

Now that this paper has reviewed examples of available data sets for research on gait recognition, and methods that may be applied to these data sets to enable such research, this section will explore promising directions for gait recognition research by outlining an example of a research project that might be conducted within each of the four categories of normal-non-longitudinal, normal-longitudinal, pathological-non-longitudinal, and pathological-longitudinal data. 

\subsection{Normal Gait and Non-Longitudinal Target}

A potential research study within this avenue of inquiry could be a gradient-definition of ‘fatigue’ through a progressive\footnote{A progressive model is a model consisting of ordered states that represent an evolution of some concept. In this case, fatigue is modeled such that the initial state is normal, while the final state is completely fatigued. And there is some qualification of states in between normal and fatigued.} model. Fatigue in this case refers to muscular depletion that renders biomechanical movement unduly difficult, which is treated as a terminal state. Such measurement would require a quantitative scale for states of fatigue that is objectively measurable through gait. There have been studies that identify fatigue by requiring participants to exercise until they achieve a state of perceived muscular depletion (Janssen, Schollhorn, Newell, Jager, Rost, Vehof, 2011). The hypothesis is that gait perturbation is a proxy indicative of fatigue by which a classifier might be trainable for attributing such perturbations to fatigued-state transitions.
\par
In order to design a model that can view the progression of fatigue, there must be a representation of an ordered sequence of fatigued states. Given the methods reviewed in this paper, Hidden Markov Models are the most intuitive option to explore. A HMM gait-cycle could be generated for each observed or measurable fatigued-state-transition, and detecting the current fatigued-state of an individual by cyclical comparison through the HMM should be sufficient to quantify their state. Other architectures for exploration of condition-based gait signals associated with fatigue also include AutoEncoders, as well as Discrete Fourier Transform of accelerometer vibrations (i.e. tremors).
\par
The PhysioBank Long Term Dynamics dataset contains information regarding healthy individuals moving constantly at different intensities for a full hour. This dataset could be sufficient for exploring the described fatigued-state-model, as the participants’ gait could reflect symptoms of fatigue.

\subsection{Normal Gait and Longitudinal Target}

The future research relevant to this section could build upon the methods defined in Section 5.1. Given a fatigue-model that quantifies the progressive states of fatigue, it is reasonable to believe that forecasting a transition to higher fatigued states is possible. This specific approach could be combined with the field of injury prevention by computing a risk factor based on the speed of transitions between fatigued states, yielding an injury-risk model based on the hypothesis that sudden and/or acute fatigued-state transitions increase the likelihood for injury.
\par
Given the wide array of features and complexity of the problem, it is likely that the best course of action is to develop some deep learning architecture utilizing the different HMMs described in Section 5.1 to learn about the transitions between states and the fluctuation of gait that ensues depending on said state. While none of the datasets reviewed contains exhausting behavior along with information on injury, the PhysioBank dataset for Long Term Movement is longitudinal and has information over multiple days including potential falls of elderly participants. It is reasonable to hypothesize that these falls are related to exhaustion in some form or another, and hence the dataset could be sufficient for initial exploration of this research opportunity by using the methods described to assess exhaustion prior to the fall with the gait-data in proximity of the fall as the chosen target variable.

\subsection{Pathological Gait and Non-Longitudinal Target}

A promising project within this category is the classification of a set of predefined pathologies. Given a training set of gait-related data where pathological disorders are a feature (the PhysioBank Neurodegenerative Diseases dataset, or the Kaggle PD Dataset), it is not difficult to extrapolate from this review that both a K-Nearest-Neighbors classifier and a Support Vector Machine could be adequate classifiers for this problem if DTW techniques are employed for comparison. 
\par
The databases described, however, contain pathologies that are known to drastically alter gait. It would be interesting to consider other data that is available for disorders, such as mania and depression as well as inflammatory responses to infection (which results in pathological gait symptomatic of fatigue), that may have distinct patterns of gait to use for a classification model of these disorders. This array of disorders would make such a project more comprehensive and informative, as a full spectrum of gait fluctuation would be surveyed, rather than just extreme fluctuation.

\subsection{Pathological Gait and Longitudinal Target}

A research pursuit within this area could follow the progressive models from Sections 5.1 and 5.2, and extend the concepts onto progressive pathologies. Certain pathological conditions, such as having legs of different lengths, are constant. However, disorders such as depression are episodic. There are time periods containing high levels of depressive symptoms and others with minimal to no depressive symptoms. The ability to forecast these disabling episodes by analyzing the progression of gait over time using data related to previous episodes using the techniques described above is a promising project. Given that these episodes could be described as anomalies, this project falls within the area of anomaly detection within time series analysis. According to Singh (2017), the state-of-the-art approach to this type of problem would be a Neural Network with a stacked-LSTM\footnote{LSTM stands for Long Short-Term Memory, which is a specific Recurrent Neural Network structure capable of understanding long-term dependencies. It is described in detail by Singh (2017).} architecture. Hence that algorithm would appear to be a good starting point.
\par
Unfortunately, there does not appear to be publicly available data that fulfills the needs of this project, but it is highly possible that data sets with relevant information may be found in non-public domains. Other similar applications may concern neurologic disorders such as Parkinson's Disease which has a very explicit progression in five stages, and the Kaggle Competition dataset referenced in Section 2.4.1 could be sufficient for exploring this project on Parkinson’s Disease. Accordingly, this project as applied to Parkinson’s Disease could develop tools with a potential application to forecasting depressive episodes\footnote{It has been shown that there is a high correlation between depression and Parkinson’s Disease. Shrag, Jahanshahi, and Quinn (2001) displayed this as well as demonstrated that the severity of depression increases with the severity of PD. Given this correlation, the tools that assist in identifying the progression of Parkinson’s Disease are likely to be helpful in approaching a similar problem with subjects diagnosed with depression.} once accompanied by satisfactory data. 

\subsection{Further Exploration of Gait Recognition for Syndromic Surveillance}

This diagnostic research agenda is also conceptually relevant for other initiatives, either pending, under development, or in progress (e.g. PTSD rehabilitation). Furthermore, it complements various clinical and
commercial use-case applications within safety, health and wellness subdomains (e.g. first-responders and high-risk industrial environments, workers compensation, geriatric care/assisted living, disability, accidental death and dismemberment). By way of general illustration, as recently as 2019, the Office of Naval Research solicited resources and
capability for basic and applied research into machine learning related to syndromic surveillance predictive adaptations to support human performance and injury prevention\footnote{See ONR BAA Special Notice N00014-19-S-SN08}. This research solicitation encompassed the intensive physical demands in often challenging environments for field support personnel, often ranging from short-term high intensity load activities to more prolonged \textit{longer-duration} activities, which might also involve sleep-cycle disruption or sleep deprivation, further contributing to fatigue and other environmental, physiological, and psychological stressors.
\par
As observed throughout the current COVID pandemic crisis, these are conditions characteristic of challenges consistently encountered by both first responders and other essential workers. Preventive measures and interventions might become more actionable, assisted by reliably adaptive diagnostics focused on recognition of pathological gait, further conditioned on detection and analysis of other corresponding stress and fatigue indicators, both external (e.g. voice stress analysis) and internal (e.g. galvanic skin response, oxidative pulse rate, heart rate variability), as well as ambient environmental
factors (e.g. temperature, humidity). The joint conditionality associated with the complexities of such syndromic surveillance and diagnostics necessitates experimental design which requires data fusion to infer indicative changes in gait associated with corresponding changes across these modalities. 
\par
We further envision the prospect of deploying and testing other sensing modalities (e.g. UWB and terahertz Radar, sonar, time-reversal signal processing) for gait recognition beyond wearable accelerometry or visual imagery (e.g., RGB, NIR/IR). We contemplate, at least during the calibration process, systematically incorporating some form of balance-sensing devices into the clinical assessments as an additional data gathering and assessment tool, in conjunction with other sensory technologies (e.g. MS Kinect, radar), for cross-validation and calibration of syndromic results. 

By way of another related illustration, the primary theme proposed by a less recent, although ongoing DARPA research initiative\footnote{See \href{https://research.usc.edu/files/2017/05/HR001117S0032-I2O-WASH-002.pdf}{https://research.usc.edu/files/2017/05/HR001117S0032-I2O-WASH-002.pdf}}
is the use of data collected from cellphone sensors to enable novel algorithms that conduct passive, on-line, and preferably real-time assessment of the health and performance conditions, by extracting latent physiological signals – which may be weak and noisy – within data to be obtained through existing mobile device sensors (e.g., accelerometer, screen, microphone). The stated objective of such extraction and analysis would be to diagnose current health status and identify latent or developing health disorders. The stated intention of the DARPA project is to develop algorithms and techniques for identifying both known indicators of physiological impairment (such as disease, illness, and/or injury) and deviations from the micro-behaviors that might be indicative of such impairment. It is also expected that additional \textit{digital biomarkers} of physiological problems could be identified during such research through the combination of “big data analytics” and medical ground truth with the purpose of condition-based health intervention.

Enablement of the fundamental viability for such condition-based intervention diagnostics relies upon the development and deployment of a mobile application that passively assesses health condition readiness (both immediately and over time), in order to provide, "(a) clinicians with plausible health conditions supported by the analysis, determination, and fusion of digital biomarkers for symptomatic indicators of corresponding disease incidence against ground truth; (b) team leaders with team
readiness information, both at the current time and in the near future; and (c) users of the personal device with information about their current state and early indicators of emergent medical conditions." 

Aside from a mobile application that passively collects and delivers relevant user-sensing data to a shared longitudinal data repository, such a program could involve enlisting existing or new medical cohorts, managing interactions with the cohort coordinators, and ensuring compliance with security and privacy requirements of said cohorts, to provide suitably anonymized and aggregated data access under appropriate protections – via an API or other such utility – to the relevant smartphone sensor generated data collected from participating cohort subjects. Implementing such a clinical research protocol entails foundational technology that for, "(a) passive continuous collection of reliable physiological data from a smartphone; (b) data analysis for identifying reliable digital biomarkers from such data; and (c) the fusion of such digital biomarkers for achieving disease identification, recognition of symptomatic precursors, as well as cognitive and physiological state of readiness assessment." The technical approach draws upon fields that include statistical data analytics (with an anticipated emphasis on signal analysis and machine learning), behavioral biometrics, and epidemiological domains \footnote{Broader issues of ethics, policy, and legality regarding human experimentation, privacy (e.g. HIPAA, GDPR), and algorithmic bias, although outside the scope of this paper are nonetheless relevant to gait recognition, syndromic surveillance and bioinformatics. By way of background, as research has become formalized, the academic community has developed formal definitions of human subject research, associated with the notion of informed consent, largely in response to abuses of human subjects. Human subject research, either medical or non-medical (e.g., social science) research involves systematic investigation with both collection and analysis of data, in order to answer a specific question. Medical research on human subjects might also involve behavioral analysis. The US Department of Health and Human Services (DHHS) defines a human research subject as a living individual about whom a research investigator – either a professional or a student – obtains data through either intervention or interaction with the individual, or via identifiable private information (32 C.F.R. 219.102(f)). As defined by HHS regulations: Intervention refers to physical procedures by which data is gathered and the manipulation of the subject and/or their environment for research purposes [45 C.F.R. 46.102(f)]. Interaction refers to communication or interpersonal contact between investigator and subject [45 C.F.R. 46.102(f)]). Private Information refers to information about behavior that occurs in a context in which an individual can reasonably expect that no observation or recording is taking place, and information which has been provided for specific purposes by an individual and which the individual can reasonably expect will not be made public [45 C.F.R. 46.102(f)] )]. Identifiable information refers to specific information that can be used to identify an individual. In 2010, the National Institute of Justice in the United States published recommended rights of human subjects: voluntary, informed consent; respect for persons treated as autonomous agents; the right to end participation in research at any time; right to safeguard integrity; benefits should outweigh cost; protection from physical, mental and emotional harm; access to information regarding research; protection of privacy and well-being. The Common Rule, first published in 1991 (aka the Federal Policy for the Protection of Human Subjects), dictated by the Office of Human Research Protections under the United States Department of Health and Human Services and serving 
as a set of guidelines for institutional review boards (IRBs), obtaining informed consent, and Assurances of Compliance for human subject participants in research studies – to which was a final rule added to the Federal Register on January 19, 2017 with an official effective date of July 19, 2018. See Human Subject and Privacy Protection, 
National Institute of Justice; "Federal Policy for the Protection of Human Subjects 'Common Rule". HHS.gov; "Federal Policy for the Protection of Human Subjects". Federal Register; "Revised Common Rule". HHS.gov. ... According to federal regulations, defining features of human subject research are that the researchers interact directly with the subject or obtain identifiable private information about the subject. Research involving digital technologies such as [syndromic surveillance] may or may not meet this definition. The institutional review board (IRB) of a research institution is typically responsible for reviewing potential research on human subjects. However, IRB directives at some institutions may be ambiguous or outdated regarding certain digital forensics, informatics and data practices.}.

\section{Conclusion}

This paper reviewed the literature and implementation of gait recognition within the four categories described by the permutations of normal versus pathological gait and longitudinal versus non-longitudinal classification for both biometric and video data. This implementation was detailed first by determining ideal preprocessing techniques on the data, followed by an in-depth survey of popular classification algorithms. Furthermore, this paper presented potential research directions in each of these four categories regarding gait recognition as a classification problem, as well as syndromic surveillance and inferences related to fatigue and other observable indicators. Inferences to normal versus pathological gait patterns can be attributed, based on both longitudinal and non-longitudinal forms of classification. A broader scope of longitudinal analysis regarding the perturbations in gait across states (i.e. physiological, emotive, and/or cognitive states) may indicate promising research directions and experimental designs, such as creating algorithmic metrics for the quantification of fatigue, or models for forecasting episodic disorders. Lastly, in conjunction with other measurements of physiological and environmental conditions, pathological gait classification might be applicable to inference for syndromic surveillance of infectious disease states or cognitive impairment.

\section{Acknowledgements}

The authors acknowledge Michael Coggins for helpful comments to an early draft of the paper. The corresponding author especially acknowledges Dennis Shasha and Ken Perlin for advice and resources. This research was supported by the RiskEcon® Lab for Decision Metrics @ Courant Institute of Mathematical Sciences NYU.





\section{References}

\begin{hangparas}{.25in}{1}
Bai J, Di C, Xiao L, Evenson KR, LaCroix AZ, et al. (2016) An Activity Index for Raw
Accelerometry Data and Its Comparison with Other Activity Metrics. 
PLOS ONE 11(8): e0160644.\ 

Benezeth, Y., Jodoin, P., Laurent, H., \& Rosenberger, C. (2010). Comparative study of 
background subtraction algorithms. Journal of Electronic Imaging,19(3), 033003. \ 

Bennabi, D., Vandel, P., Papaxanthis, C., Pozzo, T., \& Haffen, E. (2013). Psychomotor Retardation in Depression: A Systematic Review of Diagnostic, Pathophysiologic, and Therapeutic Implications. BioMed Research International, 2013, 1-18. \ 

Boser, B. E., Guyon, I. M., \& Vapnik, V. N. (1992). A training algorithm for optimal margin   classifiers. Proceedings of the Fifth Annual Workshop on Computational Learning Theory - COLT 92. \ 

Bouchrika, I., \& Nixon, M. S. (n.d.). Model-Based Feature Extraction for Gait Analysis and 
Recognition. Computer Vision/Computer Graphics Collaboration Techniques Lecture Notes in Computer Science, 150-160.\ 

Boulgouris, N.V. \& Hatzinakos, Dimitrios \& Plataniotis, Konstantinos. (2005). Gait recognition: A challening signal processing technology for biometric identification. Signal Processing Magazine, IEEE. 22. 78 - 90. 10.1109/MSP.2005.1550191. 

Burges, C. J. (1998). A Tutorial on Support Vector Machines for Pattern Recognition. Data   Mining and Knowledge Discovery,2(2), 121-167. \ 

Chandola, V., Banerjee, A., \& Kumar, V. (2009). Anomaly detection. ACM Computing
Surveys,41(3), 1-58. \ 

Cortes, C., \& Vapnik, V. (1995). Support-vector networks. Machine Learning,20(3), 273-297.\ 

Csáji, B. C. (2012) Approximation with artificial neural networks, M.S. thesis, Dept. Science, Eötvös Loránd Univ., Budapest, Hungary, 2001.\ 

Ekinci, M., \& Aykut, M. (2007). Human Gait Recognition Based on Kernel PCA Using
Projections. Journal of Computer Science and Technology,22(6), 867-876.\

Gadaleta, M., \& Rossi, M. (2018). IDNet: Smartphone-based gait recognition with convolutional neural networks. Pattern Recognition,74, 25-37.\ 

Goldberger, A., Amaral, L., Glass, L., Hausdorff, J., Ivanov, P. C., Mark, R., ... \& Stanley, H. E. (2000). PhysioBank, PhysioToolkit, and PhysioNet: Components of a new research resource for complex physiologic signals. Circulation [Online]. 101 (23), pp. e215–e220.\

Gross, R. \& Shi, J. (2001) The CMU Motion of Body (MoBo) Database. Robotics Institute at Carnegie Mellon University.\ 

Gudmundsson, S., T. P. Runarsson and S. Sigurdsson, "Support vector machines and dynamic time warping for time series," 2008 IEEE International Joint Conference on Neural Networks (IEEE World Congress on Computational Intelligence), Hong Kong, 2008, pp. 2772-2776. \ 

Hai, H., Thuc, H. (2015). Cyclic HMM-Based Method for Pathological Gait Recognition from   Side View Gait Video. International Journal of Advanced Research in Computer Engineering \& Technology, vol. 4, Issue 5, May 2015.\ 

Iwana, B. K., Frinken, V., \& Uchida, S. (2016). A Robust Dissimilarity-Based Neural Network  for Temporal Pattern Recognition. 2016 15th International Conference on Frontiers in   Handwriting Recognition (ICFHR). \ 

Janssen, D., Schöllhorn, W. I., Newell, K. M., Jäger, J. M., Rost, F., \& Vehof, K. (2011).   Diagnosing fatigue in gait patterns by support vector machines and self-organizing maps.   Human Movement Science, 30(5), 966-975.\ 

Jiliang Tang, Salem Alelyani, and Huan Liu. 2014a. Feature selection for classification: a review. Data Classification: Algorithms and Applications (2014), 37. 

Ju Man and Bir Bhanu, "Individual recognition using gait energy image," in IEEE Transactions
on Pattern Analysis and Machine Intelligence, vol. 28, no. 2, pp. 316-322, Feb. 2006. \ 

Khokhlova, Margarita \& Migniot, Cyrille \& Morozov, Alexey \& Sushkova, Olga \& Dipanda, Albert. (2019). Normal and Pathological Gait Classification LSTM model. Artificial Intelligence in Medicine. 94. 10.1016/j.artmed.2018.12.007. 

Krizhevsky, A., Sutskever, I., and Hinton, G. E. (2012) ImageNet classification with deep    convolutional neural networks. In NIPS, pp. 1106–1114, 2012.\ 

Kumar, S. K., Kant, L. K., \& Shachi, S. (2013). HMM Based Enhanced Dynamic Time Warping   Model for Efficient Hindi Language Speech Recognition System. Mobile Communication and Power Engineering Communications in Computer and Information Science, 200-206.\ 

LeCun, Yann; Léon Bottou; Yoshua Bengio; Patrick Haffner (1998). Gradient-based learning   applied to document recognition. Proceedings of the IEEE. 86 (11): 2278–2324. \ 

Rumelhart, D. E., Hinton, G. E., \& Williams, R. J. (1986). Learning representations by       back-propagating errors. Nature,323(6088), 533-536.\ 

Sarkar, S., P. Jonathon Phillips, Z. Liu, I. Robledo, P. Grother, K. W. Bowyer, “The Human ID Gait Challenge Problem: Data Sets, Performance, and Analysis,” IEEE Transactions on Pattern Analysis and
Machine Intelligence, vol. 27, no. 2, pp. 162 – 177, Feb. 2005.\

Saxe, R. C. (2018), Classifying the Quality of Movement via Motion Capture and Machine Learning, New York University Master's Thesis. \

Schrag A, Jahanshahi M, Quinn NP. What contributes to depression in Parkinson's disease?.   Psychol Med. 2001;31(1):65-73.\ 

Shaikh, S. H., Saeed, K., \& Chaki, N. (2014). Gait recognition using partial silhouette-based
approach. 2014 International Conference on Signal Processing and Integrated Networks. \ 

Shirke, S., Pawar, S., \& Shah, K. (2014). Literature Review: Model Free Human Gait 
Recognition. 2014 Fourth International Conference on Communication Systems and Network Technologies.\ 

Singh, A. (2017) Anomaly Detection for Temporal Data Using Long Short-Term Memory (LSTM), M.S. thesis, School of Information and Communication Technology, Kth Royal Institute of Technology, Stockholm, Sweden, 2017.\ 

Su, H., Liao, Z., \& Chen, G. (2009). A gait recognition method using L1-PCA and LDA. 2009   International Conference on Machine Learning and Cybernetics. \ 

Switonski, A., Josinski, H., Zghidi, H., \& Wojciechowski, K. (2012). Selection of pose     configuration parameters of motion capture data based on dynamic time warping.     International Journal of Computer, Electrical, Automation, Control and Information .i Engineering,6, 11th ser. \ 

Vehbi Olgac, A \& Karlik, Bekir. (2011). Performance Analysis of Various Activation Functions in Generalized MLP Architectures of Neural Networks. International Journal of Artificial Intelligence And Expert Systems, 1. 111-122. \ 

Whittle, M. W. (2002). Gait analysis: an introduction. Edinburgh: Butterworth-Heinemann. \ 
\end{hangparas}

\end{document}